\documentclass{article}
\usepackage{spconf,amsmath,graphicx, math-common}
\usepackage{caption}
\usepackage{color}
\usepackage{bm}
\usepackage{hyperref}

\title{Mixture Model Auto-Encoders:\\
Deep Clustering Through Dictionary Learning}
\name{Alexander Lin$^{\star}$ \qquad Andrew H. Song$^{\dagger}$ \qquad Demba Ba$^{\star}$}
  
  \address{$^{\star}$School of Engineering and Applied Sciences, Harvard University, Boston, MA, USA \\
      $^{\dagger}$Massachusetts Institute of Technology, Cambridge, MA, USA
     }

\begin{document}
%
\maketitle
\begin{abstract}
State-of-the-art approaches for clustering high-dimensional data utilize deep auto-encoder architectures. Many of these networks require a large number of parameters and suffer from a lack of interpretability, due to the black-box nature of the auto-encoders. We introduce \emph{Mixture Model Auto-Encoders} (MixMate), a novel architecture that clusters data by performing inference on a generative model. Built on ideas from \textit{sparse dictionary learning} and \textit{mixture models}, MixMate comprises several auto-encoders, each tasked with reconstructing data in a distinct cluster, while enforcing sparsity in the latent space. Through experiments on various image datasets, we show that MixMate achieves competitive performance versus state-of-the-art deep clustering algorithms, while using orders of magnitude fewer parameters.
\end{abstract}
\section{Introduction}
\label{sec:intro}

\emph{Clustering} is a fundamental task for dividing data into groups without supervised labels. \emph{Deep clustering} is a recent line of work that leverages deep neural networks to improve clustering for high-dimensional data, such as images.  
One line of research uses a single auto-encoder to project data into a low-dimensional space that is friendly for simple clustering algorithms~\cite{xie2016unsupervised, yang2017towards}. Another line of work employs a mixture-of-experts approach~\cite{Jacob91}, where $K$ different auto-encoders partition the dataset into $K$ parts~\cite{chazan2019deep, opochinsky2020k}. While more accurate than simple clustering algorithms, these frameworks typically have black-box architectures that lack interpretability and require a large number of parameters.      

\emph{Model-based deep learning} is an emergent methodology for marrying the interpretability of signal processing models with the learning efficiency and representational power of neural networks \cite{shlezinger2020model}.  It has shown success in various imaging applications~\cite{tolooshams2020deep, wu2019learning, tolooshams2018scalable, aggarwal2018modl}.
The model-based approach starts with a generative model of data, which enables the incorporation of domain knowledge. The unfolding of classical optimization algorithms then leads to multi-layer neural networks for conducting inference. 
This connection enables the use of deep learning tools (e.g. backpropagation, graphics processing units) for accelerated learning \cite{tolooshams2021pudle}, while maintaining the interpretability of the original model.

We introduce \emph{Mixture Model Auto-Encoders} (MixMate), a novel architecture for \emph{model-based deep clustering} of images.
It is derived from a mixture of sparse dictionary learning models, inspired by the wealth of evidence for the sparsity of natural images with respect to suitable dictionaries \cite{Aharon06, starck2015sparse}. MixMate comprises $K$ different auto-encoders with an attention module, all of which are parameterized by a set of cluster-specific dictionaries. We train the network with a loss function from the classical Expectation-Maximization algorithm \cite{dempster1977maximum}. These properties enable MixMate to achieve superior clustering performance on benchmark image datasets with far fewer parameters.  MixMate also exhibits other benefits, such as an interpretable architecture, a simpler initialization scheme, and the ability to cluster incomplete data.\footnote{Our code can be found at \href{https://github.com/al5250/mixmate}{https://github.com/al5250/mixmate}}



\section{Generative Model}
\label{sec:generative}
Given a collection of high-dimensional data, \emph{sparse dictionary learning} posits that each data point $\bd y \in \R^M$ can be represented as a sparse combination $\bd x \in \R^D$ of global atoms $\bd a_1, \bd a_2, \ldots, \bd a_D \in \R^M$ arranged as columns of a \emph{dictionary} $\bd A \in \R^{M \times D}$.  Collectively, $\bd A$ and the sparsity of $\bd x$ define a highly constrained subspace to explain $\bd y$.

If a dataset can be partitioned into $K$ natural clusters (e.g. $K=10$ digit identities in a handwritten digits dataset), it is reasonable to assume that each cluster lies in a different subspace. In this context, we assume that objects belonging to cluster $k \in \{1, \ldots, K\}$ are generated by a cluster-specific dictionary $\bd A_k$. The relative size of each cluster is encoded into the prior $\bm{\pi}=[\pi_1,\ldots,\pi_K]\in [0,1]^K$, with $\sum_{k=1}^K \pi_k = 1$. Putting these together, we employ the following \emph{sparse dictionary learning mixture model} as the generative process for each element $\bd y$ in a dataset $\mathcal{Y} = \{\bd y_1, \bd y_2, \ldots, \bd y_N\}$,
\begin{align}
z &\sim \text{Categorical}(\bm{\pi}), \label{model} \nonumber \\
\bd x &\sim \text{Laplace}(\lambda) \propto \exp(-\lambda \norm{\bd x}_1),  \nonumber\\
\bd y \given \bd x, z = k &\sim \mathcal{N}(\bd A_k \bd x, \bd I) \propto \exp(-\norm{\bd y - \bd A_k \bd x}_2^2).
\end{align}
The Laplace prior on $\bd x$, which induces the $\ell_1$ norm on $\bd x$, constrains $\bd y$ to be constructed from only a few columns of $\bd A_k$, with $\lambda$ dictating the strength of the sparsity penalty.

\subsection{Parameter Estimation \& Inference for Clustering}
With the generative model, we can cluster data by inferring the cluster identity $z$ for each $\bd y$ through the posterior $p(z|\bd y)$. This requires estimation of the latent code $\bd x$ and the model parameters $\theta = \{\bd A_1, \ldots, \bd A_K, \bm{\pi}\}$. One effective method for accomplishing this goal is the expectation-maximization (EM) algorithm~\cite{dempster1977maximum}. 
EM fits the model to $\mathcal{Y}$ by alternating between an E-Step and an M-Step, by minimizing the following loss function $L(\theta)$ based on Eq. \eqref{model} until convergence:
\begin{align}
&L(\theta) = -\E_{p(\bd x, z \given \bd y)}[\log p(\bd x, z, \bd y)] \label{em} \\
&= -\sum_{k=1}^K \underbrace{p(z = k \given \bd y)}_{\emph{Attention}} \cdot \underbrace{\mathbb{E}_{p(\bd x \given \bd y, z =k)}}_{\emph{Encoding}}[\underbrace{\log p(\bd x, z = k, \bd y )}_{\emph{Decoding}}]. \nonumber
\end{align}
 The \emph{E-Step} enables computation of $L(\theta)$ by solving for latent variables $\{\bd x, z\}$ given a current parameter estimate $\tilde{\theta}$, and the \emph{M-Step} updates the model parameters to a new estimate $\tilde{\theta}^\text{new}$. 
In the next section, we show how we can use the model and the loss function to derive a deep clustering network.

\section{Mixture Model Auto-Encoders}\label{sec:arx}
\begin{figure}
\begin{center}
\includegraphics[scale=0.18]{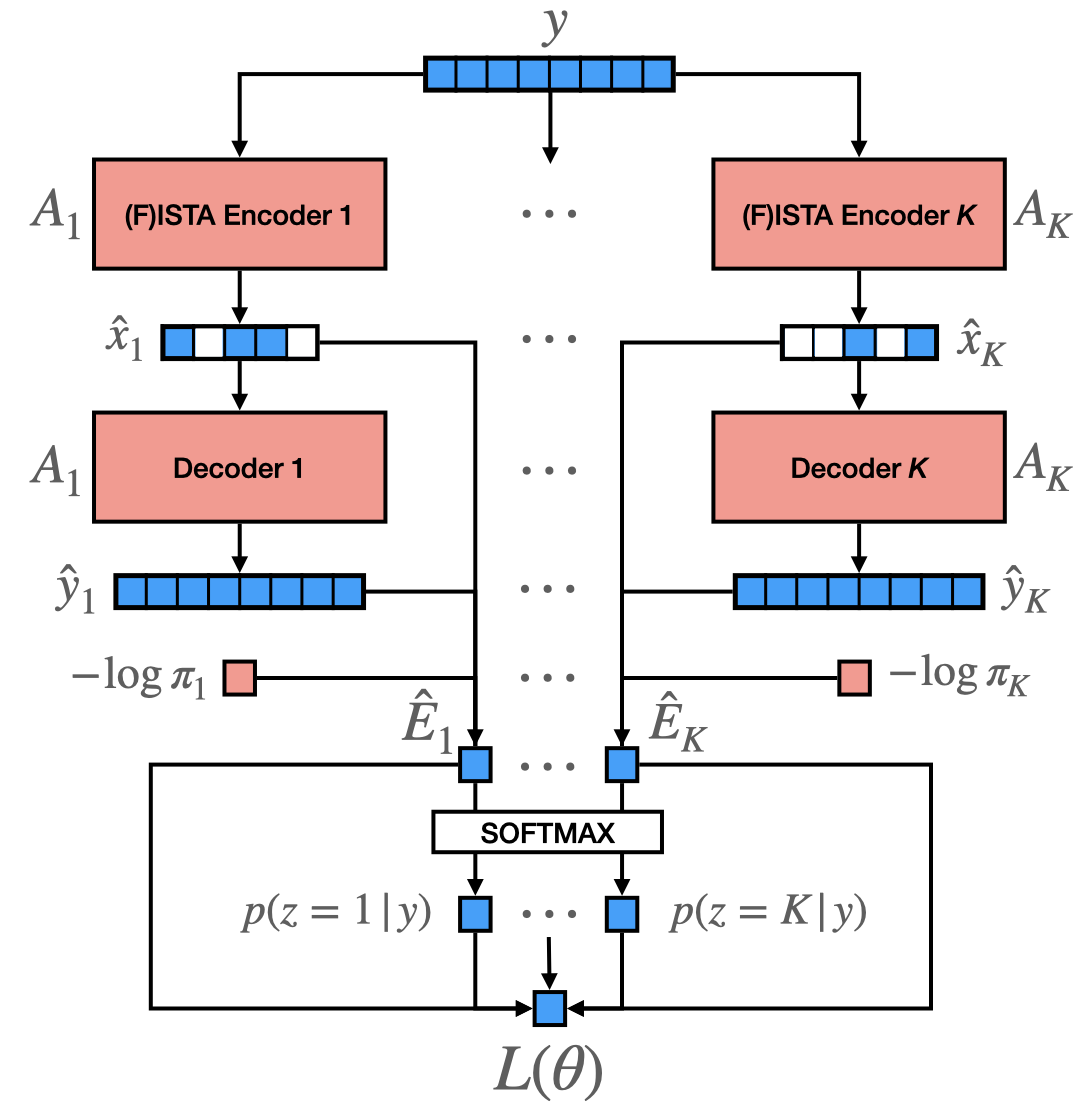}
\caption{A diagram of the MixMate architecture.  Arrows outline the flow of information for the \emph{forward} pass.  Note the tying of weights between the encoders and decoders.} \label{fig:diagram}
\end{center}
\vspace{-1.5em}
\end{figure}

We now introduce our clustering framework, \textit{Mixture Model Auto-Encoders (MixMate)}. MixMate (depicted in Fig. \ref{fig:diagram}) has three main parts -- the encoding, decoding, and attention modules -- each of which corresponds to a key term in Eq. \eqref{em}. 

\subsection{Architecture} \label{sec:arch}
\textbf{Encoder} We observe that $p(\bd x \given \bd y, z = k)$ acts as an \emph{encoder} that maps the data $\bd y$ to a posterior distribution over the sparse code $\bd x$ using the $k$-th cluster's parameters. However, this posterior is not analytically tractable, due to non-conjugacy between the Laplace prior and the Gaussian likelihood, preventing the computation of the expectation in Eq. \eqref{em}.
To resolve this issue, we approximate the posterior by its \textit{mode} $\hat{\bd x}_k$ \cite{tolooshams2020deep},
\begin{align}
&\hat{\bd x}_{k} =  \arg \min_{\bd x} \norm{\bd y - \bd A_k \bd x}_2^2 + \lambda \norm{\bd x}_1. \label{encode}
\end{align}
This choice has two consequences. First, it reduces the expectation of Eq. \eqref{em} to evaluation of a single term at $\hat{\bd x}_k$. Second, it turns posterior inference into a \emph{sparse coding} problem (Eq. \eqref{encode}). We run the fast iterative shrinkage-thresholding algorithm (FISTA)~\cite{Beck09, gregor2010learning} for $L$ iterations to obtain $\hat{\bd x}_k = \bd \alpha_{(L)}$, via the following recurrence indexed by $\ell = 1, \ldots, L$:
\begin{align}
t_{(0)} = 1, \quad &t_{(\ell)} = \tfrac{1}{2} (1 + \sqrt{1 + 4t_{(\ell - 1)}^2}), \nonumber \\
\bd \alpha_{(0)} = \bd 0, \quad &\bd \alpha_{(\ell)} = f_{\eta \lambda}(\bd \beta_{(\ell - 1)} + \eta \bd A_k^\top (\bd y - \bd A_k \bd \beta_{(\ell - 1)})), \nonumber \\
\bd \beta_{(0)} = \bd 0, \quad &\bd \beta_{(\ell)} = \bd \alpha_{(\ell)} + \tfrac{t_{(\ell - 1)} - 1}{t_{(\ell)}}(\bd \alpha_{(\ell)} - \bd \alpha_{(\ell - 1)}).
\label{ista}
\end{align}
Here, ${\bd 0} \in \R^D$ is the zero vector, $\eta$ is the step size, and $f_\gamma: \R^D \to \R^D$ is the soft-thresholding operator (the double-sided rectified linear unit) with threshold $\gamma$.  Eq. \eqref{ista} specifies a cascade of linear and nonlinear operations.  Thus, it can be interpreted as an $L$-layer neural network encoder with parameters $\bd A_k$ that inputs $\bd y$ and outputs $\bhat x_k$.  
As there are $K$ different dictionaries, the encoding module contains $K$ parallel FISTA encoders, each returning a different code $\hat{\bd x}_{k}$ (Fig. \ref{fig:diagram}).

\noindent \textbf{Decoder} We now use $\hat{\bd x}_{k}$ to evaluate $\log p(\bd x, z =k , \bd y )$.  We call the negation of this quantity the $k$-th \emph{energy} $\hat{E}_k$ of $\bd y$,
\begin{align}
 \hat{E}_k &=  -[ \log p(\bd y \given \hat{\bd x}_k, z = k) + \log p(\hat{\bd x}_k) + \log p(z = k)] \nonumber \\
 &= \underbrace{\norm{\bd y - \bd A_k \hat{\bd x}_k}_2^2}_{\emph{Reconstruction}} + \underbrace{\lambda \norm{\hat{\bd x}_k}_1}_{\emph{Regularization}} - \underbrace{\log \pi_k}_{\emph{Bias}}. \label{energy}
\end{align}

Observe that $\hat{E}_1, \ldots, \hat{E}_k$ are functions of the input data $\bd y$ and the model parameters $\theta = \{\bd A_1, \ldots, \bd A_K, \pi\}$. In Eq. \eqref{energy}, the first term involves passing the sparse code $\hat{\bd x}_k$ through a \emph{decoder} $\bd A_k$ to obtain the $k$-th cluster's reconstruction of the data $\hat{\bd y}_k = \bd A_k \hat{\bd x}_k$.  
Therefore, Eqs. \eqref{encode} and \eqref{energy} specify $K$ parallel \emph{auto-encoders} that map $\bd y$ to $K$ different reconstructions.  The two other terms capture the sparsity of $\bhat{x}_k$ and the bias of cluster $k$, which are also used to determine the cluster of $\bd y$.



\noindent \textbf{Attention} The component of Eq. \eqref{em} that is last computed by the network (Fig. \ref{fig:diagram}) is
\begin{align}
&p(z = k \given \bd y) 
\propto \pi_k \cdot \int p(\bd x) p(\bd y \given \bd x, z = k)d \bd x \nonumber \\
&\approx \pi_k \cdot \left\{\max_{\bd x} p(\bd x) \cdot p(\bd y \given \bd x, z =k) \right\} = \exp(-\hat{E}_k),\label{attn}
\end{align}
where we have again dealt with non-conjugacy in the integral using a mode approximation. Normalizing Eq. \eqref{attn} yields
\begin{align}
p(z = k \given \bd y) \approx \frac{\exp(-\hat{E}_k)}{\sum_{k'=1}^K \exp(-\hat{E}_{k'})}, \label{prob}
\end{align}
which is equivalent to the \emph{softmax} nonlinearity $\bd \sigma : \R^K \to \R^K$ applied to the negation of $\hat{\bd E} = [\hat{E}_1, \ldots, \hat{E}_K]$. Therefore, $\hat{E}_k$ measures how likely $\bd y$ belongs to the $k$-th cluster, with a lower value indicating higher likelihood. Consequently, the loss function of MixMate in Eq.~\eqref{em} can be expressed as
\begin{align}
L(\theta) \approx \hat{\bd E}^\top \bd \sigma(-\hat{\bd E}), \label{loss}
\end{align}
which reflects a form of \textit{attention}~\cite{vaswani2017attention}. Eq. \eqref{loss} shows that the goal of MixMate is to minimize a weighted average of the energies across the $K$ clusters.  To gain some intuition for why this objective is suited for clustering data, consider a sample $\bd y$ with $\hat{\bd E}$ that is small in the $k$-th component, but large for all other components.  Thus, $\bd \sigma(-\hat{\bd E}) \approx \bd e_k$ where $\bd e_k$ is the $k$-th unit vector, suggesting that MixMate is certain that $\bd y$ belongs to cluster $k$.  The resulting loss $L(\theta) \approx \hat{E}_k$ implies that MixMate will focus on lowering the energy of cluster $k$ for $\bd y$ and ignore the other clusters.  Therefore, as MixMate is trained, each auto-encoder will attenuate to a different portion of the dataset, thereby inducing a natural clustering of the data.

\subsection{Training and Initialization} \label{sec:init}

\textbf{Training} As a typical deep learning framework, MixMate alternates between a \emph{forward pass} and a \emph{backward pass} over several epochs.  Each step involves a mini-batch $\mathcal{B} \subseteq \mathcal{Y}$ of data points. The \emph{forward pass} passes each $\bd y_i \in \mathcal{B}$ through MixMate to compute $p(z_i|\bd y_i)$ and $L_i(\theta)$, corresponding to the E-Step.  The \emph{backward pass} updates the parameters to $\tilde{\theta}^{\text{new}}$ via \textit{backpropagation} of the mini-batch loss $\frac{1}{|\mathcal{B}|} \sum_{i} L_i(\theta)$ through the architecture, corresponding to the M-Step.  After training, a forward pass infers the assigned cluster for each data point.   

\noindent\textbf{Initialization} Prior works have dedicated substantial efforts to network initialization to ensure strong clustering performance.
The majority of these require \emph{pre-training} for the sub-components of the architecture~\cite{xie2016unsupervised, yang2017towards, chazan2019deep, opochinsky2020k}. Although effective, pre-training complicates the application of clustering algorithms, since it requires a non-trivial amount of 1) computation time and 2) effort in tuning the hyperparameters.

For MixMate, we introduce a simple initialization procedure that avoids pre-training.  We select $D$ data points from the training set assumed to belong to cluster $k$ as the initial columns of $\bd A_k$. This assumes that a data point is reasonably modeled as a linear combination of other points in the \textit{same} cluster~\cite{elhamifar2013sparse, tankala2021kdeep}. 
These points are determined by applying an off-the-shelf clustering algorithm on a sampled subset  $\mathcal{S} \subseteq \mathcal{Y}$ such that $|\mathcal{S}| \ll N$ to obtain partitions $\mathcal{S}_1, \ldots, \mathcal{S}_K$.  Then, $D$ points in each $\mathcal{S}_k$ are used as the initial columns of $\bd A_k$. We can use algorithms that would be expensive to run on the full dataset (e.g. spectral clustering, sparse subspace clustering).       

\subsection{Comparison to Other Frameworks}
In contrast to other deep networks \cite{xie2016unsupervised, yang2017towards, chazan2019deep, opochinsky2020k}, MixMate learns and backpropagates through \emph{tied} weights between the encoding and decoding modules, as a direct consequence of the generative model. 
Thus, MixMate is often much smaller in size compared to other deep-clustering architectures. 

Within the literature, Deep Auto-Encoder Mixture Clustering (DAMIC) \cite{chazan2019deep} and $K$-Deep Auto-Encoder ($K$-DAE) \cite{opochinsky2020k} are most similar to MixMate, also using $K$ different auto-encoders. 
The key difference is that their auto-encoders are not derived from a generative model and have black-box designs that sacrifice interpretability for architectural flexibility.

\section{EXPERIMENTS}\label{sec:experiments}

\label{sec:pagestyle}

 We assessed the clustering performance of MixMate on three datasets, all with $K=10$ ground-truth clusters: (a) \textbf{MNIST}, $N = 70,000$ handwritten digits of size $28 \times 28$ ($M = 784$), (b) \textbf{FashionMNIST} \cite{xiao2017fashion}, $N = 70,000$ fashion images of size $28 \times 28$ ($M = 784$), and (c) \textbf{USPS} \cite{hull1994database}, $N = 11,000$ images of size $16 \times 16$ ($M = 256$). We evaluated performance using three metrics \cite{chazan2019deep, opochinsky2020k} -- (a) normalized mutual information (\textbf{NMI}), (b) adjusted Rand index (\textbf{ARI}), and (c) clustering accuracy (\textbf{ACC}).  Since MixMate returns soft cluster assignments, we took the maximum probability (lowest energy cluster) for each data point as its deterministic assignment. 

In all experiments, we used MixMate with $K = 10$.  We set $D=50$ per auto-encoder for MNIST and FashionMNIST, and $D=30$ for USPS, due to smaller dataset size. Each encoder ran the FISTA algorithm for $L = 15$ iterations with step size $\eta = 0.04$. For sparsity, MNIST and FashionMNIST used $\lambda = 0.75$, while USPS used $\lambda = 0.25$ since its images are 3$\times$ smaller. Since the datasets we consider are known to contain balanced classes, we fixed $\pi_k = 1/K$ for all $k$, though in general, our framework allows for these $\pi_k$ to be learned. We applied sparse subspace clustering (SSC) with default parameters \cite{elhamifar2013sparse} to $|\mathcal{S}| = 2000$ randomly selected data points, to obtain the initial dictionaries. SSC ran in fewer than 5 seconds due to the small subset size $|\mathcal{S}| \ll N$. 

We recorded the performance on the full dataset \emph{after initialization} (\textsc{init}) (i.e. before learning dictionaries)  and \emph{after training} (\textsc{train}). To be consistent with prior work \cite{opochinsky2020k}, we used the Adam optimizer \cite{kingma2014adam},  $T = 50$ epochs, a  batch size of 256 examples, and a learning rate of $0.001$. For simplicity, we used neither batch normalization nor early stopping.  

\begin{table}
\caption{Clustering metrics for various algorithms (highest numbers for each row are in \textbf{bold}). Columns marked with $*$ indicate numbers from \cite{opochinsky2020k}. The metrics are calculated as the mean over five trials.  We include ``$\pm$ standard deviation" over the five trials for the fully trained MixMate numbers.
}
\label{tab:results}
\centering
\resizebox{0.48\textwidth}{!}{
\begin{tabular}{c|cccccc}
\hline \hline
& DEC* & DCN* & DAMIC* & $K$-DAE* & \multicolumn{2}{c}{MixMate } \\ 
& & & & & \textsc{init} & \textsc{train} \\
\hline
\textbf{MNIST}&&&&&&\\
NMI & 0.80 & 0.81 & \textbf{0.86} & \textbf{0.86} & 0.75  & \textbf{0.86} $\pm$ 0.03 \\
ARI & 0.75 & 0.75 & 0.82 & 0.82 & 0.72  & \textbf{0.85} $\pm$ 0.04  \\
ACC & 0.84 & 0.83 & 0.88 & 0.88 & 0.84 & \textbf{0.92} $\pm$ 0.04 \\
Params & 2.1 M & 2.1 M & 22.1 M & 21.4 M & \multicolumn{2}{c}{0.4 M} \\
\hline
\textbf{Fashion}&&&&&&\\
NMI & 0.54 & 0.55 & 0.65 & 0.65 & 0.60  & \textbf{0.68} $\pm$ 0.02 \\
ARI & 0.40 & 0.42 & 0.48 & 0.48 & 0.44  & \textbf{0.52} $\pm$ 0.01 \\
ACC & 0.51 & 0.50 & 0.60 & 0.60 & 0.57 & \textbf{0.63} $\pm$ 0.01 \\
Params & N/A & N/A & N/A & N/A & \multicolumn{2}{c}{0.4 M} \\
\hline
\textbf{USPS}&&&&&&\\
NMI & 0.77 & 0.68 & 0.78 & 0.80 & 0.79  & \textbf{0.82} $\pm$ 0.01\\
ARI & N/A & N/A & 0.70 & 0.71 & 0.73  & \textbf{0.76} $\pm$ 0.02\\
ACC & 0.76 & 0.69 & 0.75 & 0.77 & 0.79 & \textbf{0.81} $\pm$ 0.03\\
Params & N/A & N/A  & N/A & N/A & \multicolumn{2}{c}{0.08 M} \\
\hline \hline
\end{tabular}}
\end{table}

\begin{figure}
        \includegraphics[scale=0.14]{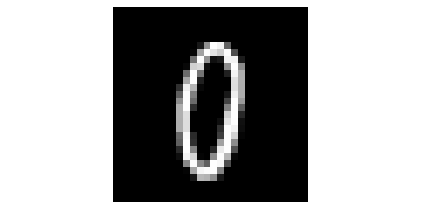}
        \includegraphics[scale=0.14]{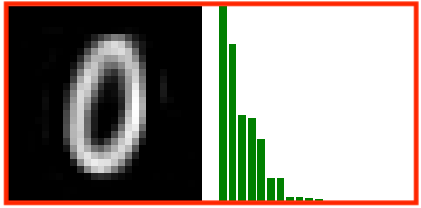}
        \includegraphics[scale=0.14]{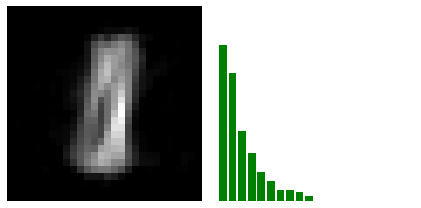} 
        \includegraphics[scale=0.14]{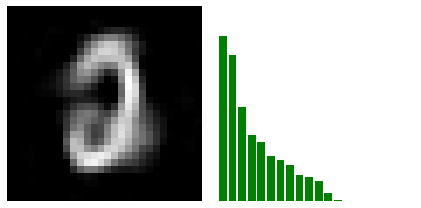} \\
        \vspace{0.5em} \hspace{1.4em} \textit{input} \hspace{2.0em} (0.025, 11) \hspace{1.0em} (0.050, 10) \hspace{1.0em} (0.033, 13) \\
        \includegraphics[scale=0.14]{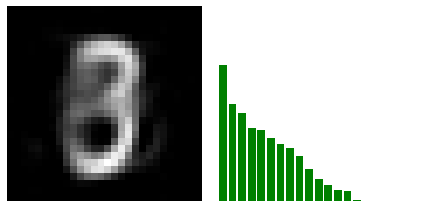}
        \includegraphics[scale=0.14]{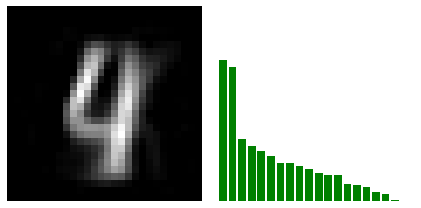}
        \includegraphics[scale=0.14]{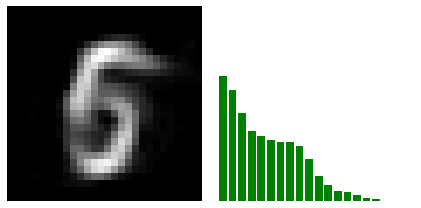} 
        \includegraphics[scale=0.14]{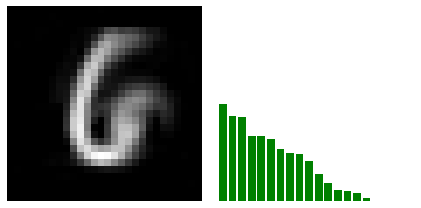}\\
        \vspace{0.5em} (0.032, 15) \hspace{1.0em} (0.031, 21) \hspace{1.0em} (0.030, 17) \hspace{1.0em} (0.036, 16)\\
        \includegraphics[scale=0.14]{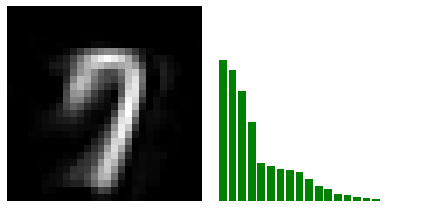} 
        \includegraphics[scale=0.14]{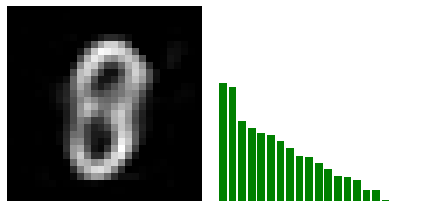} 
        \includegraphics[scale=0.14]{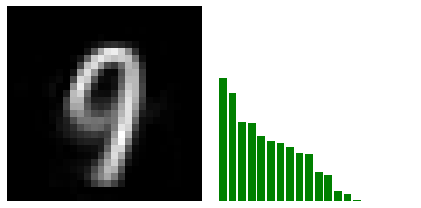} \\
        \vspace{0.5em} (0.033, 17) \hspace{1.0em} (0.025, 18) \hspace{1.0em} (0.025, 15)\\
		\captionof{figure}{Sample MNIST image $\bd y$, with MixMate's $K=10$ reconstructions $\hat{\bd y}_k$ and sparse codes $\hat{\bd x}_k$ (green).  Each image is labeled with (MSE, L0), corresponding to the mean-squared error of $\hat{\bd y}_k$ and the $\ell_0$-norm of $\hat{\bd x}_k$, respectively.  Observe that ``0", ``8", and ``9" have similar MSE, but ``0" has a sparser $\bhat{x}_k$.  MixMate clusters the digit as a ``0" (boxed in red).} \label{fig:samps}
	\end{figure}

\begin{figure}
    \centering
    \includegraphics[scale=0.35]{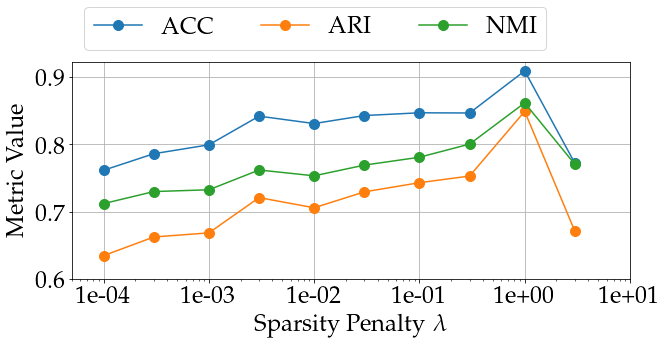}
    \caption{The metrics ACC, ARI, and NMI as a function of the sparsity penalty $\lambda$ (mean over five trials).}
    \label{fig:sparsity}
\end{figure}

\noindent \textbf{Clustering Results} 
Table \ref{tab:results} shows the results averaged over five trials.  We also included results for other unsupervised deep clustering algorithms -- Deep Embedded Clustering (DEC) \cite{xie2016unsupervised}, Deep Clustering Network (DCN) \cite{yang2017towards}, DAMIC \cite{chazan2019deep}, and $K$-DAE \cite{opochinsky2020k}. We observed that the trained MixMate almost always outperforms the baselines. It achieved superior performance while being much smaller, e.g., up to \textit{50$\times$ fewer parameters} for the MNIST dataset. The \textsc{init} column in Table \ref{tab:results} also shows that an \emph{untrained} MixMate (with SSC initialized-dictionaries and a single forward pass to cluster the dataset) is competitive with other \emph{fully trained} networks.

\noindent\textbf{The Role of the Generative Model} We discovered two patterns for how factors of $\hat{E}_k$ contributed towards clustering. In the first case, the reconstruction loss for a specific cluster was dominantly low and solely determined the cluster assignment. 
This was because each dictionary had learned columns for a particular part of the dataset (e.g. one of the digits in MNIST).
In the second case, the reconstruction losses for several clusters were similar, but the ``correct" cluster $k$ had a sparser code, leading to a lower energy $\hat{E}_k$. An example of this scenario is depicted in Fig.~\ref{fig:samps}.  This shows that both reconstruction and latent sparsity are important factors for clustering.    


\noindent \textbf{The Effect of Varying $\pmb{\lambda}$} We performed an ablation study by training MixMate on MNIST and varying $\lambda$ (Fig. \ref{fig:sparsity}).  We observed that clustering performance suffered when $\lambda$ was too small or too large.  With large $\lambda$, all auto-encoders poorly reconstructed the input, as the number and the amplitude of the latent codes were highly restricted.  With small $\lambda$, every auto-encoder reconstructed the input similarly well, as the learned dictionaries were similar to each other and thus lost cluster specificity. In both cases, the latent codes were either too sparse (large $\lambda$) or too dense (small $\lambda$) and did not play a discriminant role. Therefore, both resulted in diffuse (rather than concentrated) cluster probabilities $\bd \sigma(-\hat{\bd E})$. This suggests that the right amount of sparsity plays an important role.


\noindent \textbf{Clustering Incomplete Data} 
Since MixMate's weights parameterize a generative model, we can adapt MixMate to handle missing data by adjusting the model.   
For an incomplete image $\bd y' \in \R^m$ (where $m\leq M$) with $M - m$ missing pixels, we can modify Eq. \eqref{model} as $\bd y' \given \bd{x}, z=k\sim\mathcal{N}(\bd{\Psi}\bd{A}_k\bd x,\bd I)$, where $\bd{\Psi}\in\{0, 1\}^{m\times M}$ is a data-specific mask that extracts rows of $\bd{A}_k$ corresponding to the locations of the observed pixels of $\bd y'$. This simple change of replacing $\bd A_k$ with $\bd \Psi \bd A_k$ can be applied to the entire architecture (Sec. \ref{sec:arch}), enabling MixMate to cluster incomplete data.
We trained the network on MNIST in which 90\% of the images had 25\% of the pixels missing uniformly at random. MixMate attained 0.86 $\pm$ 0.04 NMI, 0.85 $\pm$ 0.06 ARI, and 0.92 $\pm$ 0.04 ACC across five trials -- similar to the scores for the clean data experiments. Such robustness can be useful for compressed sensing and remote sensing, where missing data is a common occurrence.

\section{Conclusion}\label{sec:conclusion}
We introduced a novel architecture for deep clustering called \emph{Mixture Model Auto-Encoders} (MixMate).  MixMate combines the interpretability of the sparse dictionary learning model with the scalability of deep learning. 
The framework of MixMate can be extended to convolutional models, non-Gaussian data distributions \cite{tolooshams2020convolutional}, as well as the integration of other priors for the latent codes and the dictionaries, such as group sparsity~\cite{theodosis2021convergence} or smoothness~\cite{song2021gaussian}.


\clearpage
\bibliographystyle{IEEEbib}
\bibliography{strings,refs}

\end{document}